\documentclass[runningheads,a4paper]{llncs}

\usepackage{amssymb}
\usepackage{graphicx}

\usepackage{url}

\begin{document}

\mainmatter  

\title{Similarity between Learning Outcomes from Course Objectives using Semantic Analysis, Bloom's taxonomy and Corpus statistics}


%
%
\author{Atish Pawar \and Vijay Mago}
%

\institute{Department of Computer Science\\Lakehead University, Thunder Bay, Canada}

%
%

\maketitle

\begin{abstract}
The course description provided by instructors is an essential
piece of information as it defines what is expected from the instructor and what he/she is going to deliver during a particular
course. One of the key components of a course description is
the Learning Objectives section. The contents of this section
are used by program managers who are tasked to compare and
match two different courses during the development of \textit{Transfer Agreements} between various institutions. This research introduces the development of semantic similarity algorithms to calculate the similarity between two learning objectives of the same domain. We present a novel methodology which deals with the semantic similarity by using a previously established algorithm and integrating it with the domain corpus utilizing domain statistics. The \textit{disambiguated} domain serves as a supervised learning data for the algorithm. We also introduce Bloom Index to calculate the similarity between action verbs in the Learning Objectives referring to the Bloom's taxonomy.
\end{abstract}

\section{Introduction}
The Learning Outcomes or Learning Objectives(LO) of a course define what the student is expected to learn by taking the course. LOs form a crucial part of any course description; hence these objectives of a course are considered as a base criterion to compare the two courses. If a student is transferring from institution A to institution B and is also attempting to transfer credits from institution A, then accurate comparison of courses is essential in deciding if the student is eligible to receive credit at institution B. This task of examining the LOs from two courses is currently completed by personnel called Program Managers/Coordinators. \newline

Program managers are also responsible for developing Transfer Program Agreements between institutes. Comparing the learning outcomes from the two course objectives is a practice followed by program managers when they are asked to compare two courses or program \cite{laanan2001transfer}. This process requires human intelligence and expertise to evaluate the course objectives. Similarly, Program Managers depend on domain experts to finalize the decision. Domain experts are persons who have knowledge of a particular field. This process depends on the human interference throughout; hence is resource and time consuming. \newline

Our intelligent system aims to automate the process of deciding whether a given student is eligible to recieve credits or not, by comparing Learning Objectives semantically. The course instructors are usually asked to follow Bloom's Taxonomy\cite{anderson2001taxonomy} when structuring the learning outcomes. Bloom’s Taxonomy provides general keyword guidelines, and a hierarchical structure to be used when defining the learning outcomes \cite{krathwohl2002revision}, see Figure 1\cite{forehand2010bloom}. But in practice, we found that instructors usually don't follow these guidelines. So, in our methodology, we limit the influence of Bloom's taxonomy by analyzing only the verbs. We use the hierarchical distribution of verbs in Bloom's taxonomy to compare learning objectives. Each layer in Bloom's taxonomy, as depicted in Figure 1, has a list of verbs associated with it \cite{verbs706bloom}. 
\begin{figure}[htbp]
\begin{center}
\includegraphics[width=0.51\textwidth, scale =1]{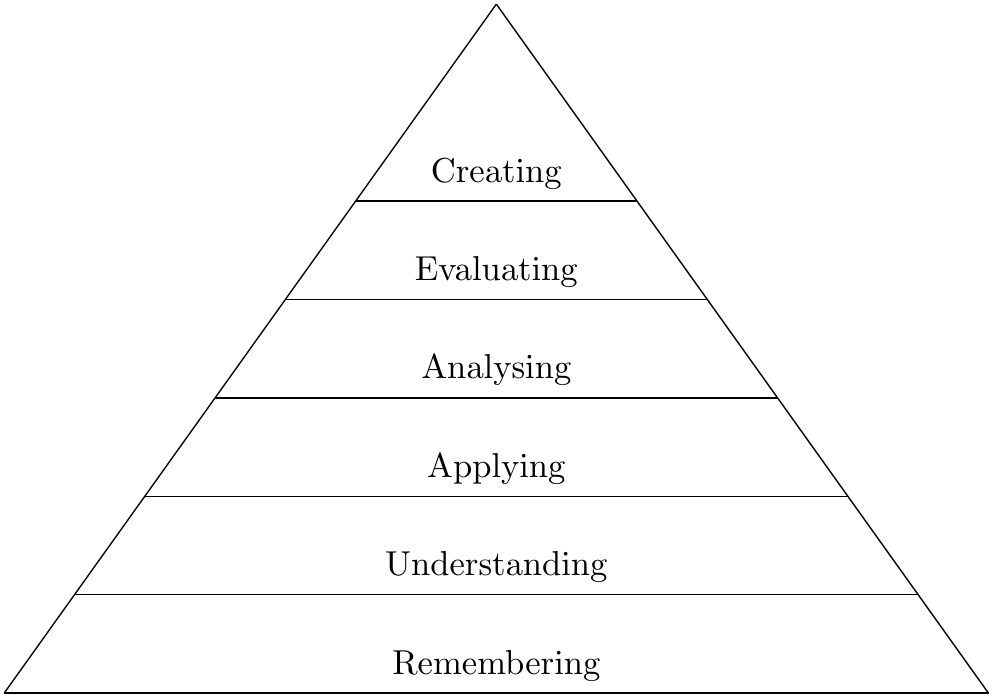}
\caption{Hierarchical Structure of Bloom's Taxonomy}
\label{Fig 1}
\end{center}
\end{figure}\\
The main contributions of this research are:
\begin{itemize}
\item Development of LO similarity measures using semantic analysis 
\item Utilizing Bloom's Taxonomy to determine the difficuly level of LOs
\item Demonstrating the effect and the usage of corpus statistics
\end{itemize}
Next section reviews some related work. Section 3 elaborates the methodology step by step. Section 4 describes the implementation in detail. Section 5 analyses the experimental results and section 6 discusses the performance of the system. Finally, section 7 explains the results in brief and draws the conclusion. 
\section{Related work}
Extensive research in the area of natural language processing has contributed valuable resolutions in the field of semantic analysis. In this section, we review some of the existing algorithms; their strengths, and weaknesses. 
The related work can be roughly classified into following sections:
\begin{itemize}
\item Similarity based on lexical databases
\item Relatedness based on web search engine results
\item Grammar-based methods
\end{itemize}
\subsubsection{Similarity based on lexical databases}
Various methods have been developed previously which use a lexical database. These methods use the hierarchical distribution of the words in the database \cite{baddeley1966short}\cite{resnik1999semantic}\cite{miller1991contextual}. Some techniques also integrate external corpus statistics with lexical database and influence the final semantic similarity \cite{li2006sentence}\cite{jiang1997semantic}. These methods have the following general limitations:
\begin{itemize}
\item The appropriate meaning of the word is not considered while computing the similarity between words which introduces inaccuracies during the earlier stage of semantic similarity calculations.
\item The corpus statistics differ for each corpus. Thus, the similarity is different for every corpus.
\item The grammar of the sentence is not considered. 
\end{itemize}
But it has following advantages:
\begin{itemize}
\item Using well-indexed lexical databases such as WordNet, have lower computational difficulties.
\item The similarity algorithm can be exploited to restrict the domain of operation. 
\end{itemize}
\subsubsection{Relatedness based on web search engine results}
This methodology uses the number of search results from an internet search engine to establish the \textit{relatedness} between words \cite{bollegala2007measuring}. This technique doesn't necessarily give the similarity between words as the number of pages indexed by the search engine are huge, and words with opposite meaning frequently occur together on the web. We have implemented the methodology to calculate the Google Similarity Distance\cite{cilibrasi2007google}, but results are not encouraging. 
\subsubsection{Grammar-based methods}
Grammar-based methods are more useful to analyze the general language sentences. Such methods ultimately depend on some measure of semantic similarity between words\cite{islam2008semantic} \cite{lee2014grammar}. 
The \textit{Sentence Text Similarity} method \cite{islam2008semantic} focuses on the semantic similarity between words as well as the \textit{String Similarity}. They also consider the order of occurrences of words. These methods work suitably for analyzing day-to-day life scenarios such as tweets, textual content from books/articles or speeches. 
When considering perfect phrases, such as LOs, the grammar remains same throughout the LOs. Hence, using such methods do not give the advantage over other methods when it comes to LOs. 

\section{Methodology}
The proposed methodology uses a semantic similarity algorithm\cite{DBLP:journals/corr/abs-1802-05667} and extends it to work with Bloom's taxonomy and corpus related to the specific domain. Figure 2 shows the modules for computing the similarity between two learning objectives.
\begin{figure}[htbp]
\begin{center}
\includegraphics[width=0.9\textwidth, scale =1]{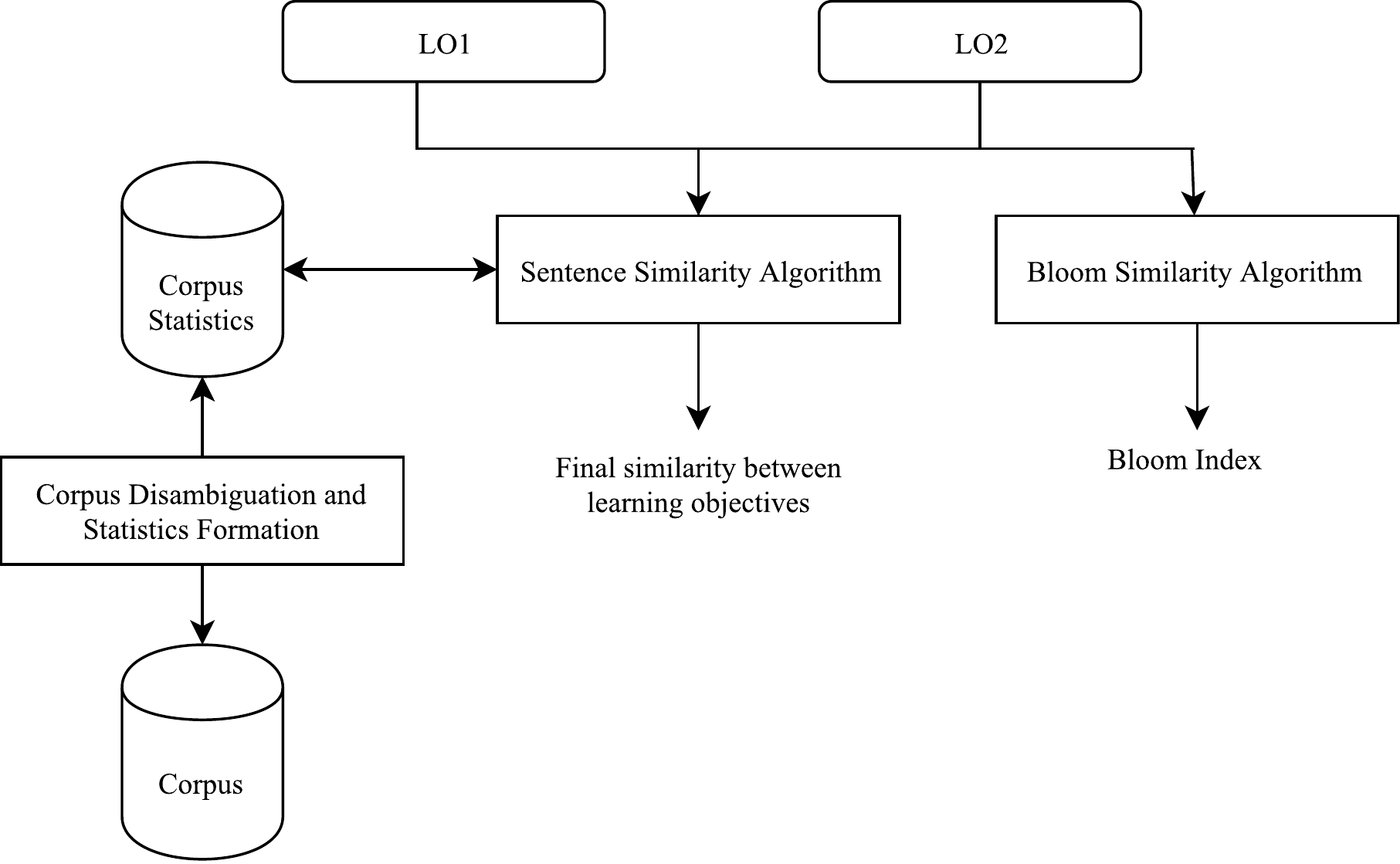}
\caption{The proposed methodology}
\label{Fig 1}
\end{center}
\end{figure}The \textit{semantic similarity algorithm} shown in Figure 2 as a process, is developed by the authors. This developed algorithm uses \textit{Synsets} from WordNet to calculate the semantic similarity between the sentences. This methodology aims to identify the correct synsets according to the meaning of the word in sentence using corpus statistics\cite{DBLP:journals/corr/abs-1802-05667}.
The methodology is divided into the following subsections:
\begin{itemize}
\item Semantic similarity algorithm
\item Bloom's taxonomy
\item Corpus statistics
\end{itemize}
\subsection{Semantic similarity algorithm in brief}
The semantic similarity algorithm used in this method is an edge-based approach which uses WordNet, a lexical database. The method to calculate the semantic similarity
between two sentences is divided into three parts:
\begin{itemize}
\item Word similarity
\item Sentence similarity
\item Word order similarity
\end{itemize}
This method first calculates the semantic similarity between words considering the meaning of the word in the context of the statement. The best result is then used to form a semantic vector for both the sentences separately. The semantic vectors formed are used to calculate the semantic similarity. The word order vector is constructed by considering the syntactic structure of the sentences, i.e., the occurrences of words concerning each other. A suite of algorithms are reported elsewhere and the interested readers of this publication can ask the authors for a copy of the paper under review.
\subsection{Bloom's taxonomy}
As discussed in section 1, a well-structured course objective describes what students will be able to learn and to do as a result of the course\cite{almerico2004bloom}. Bloom's taxonomy is well-known, established, hierarchically structured model which contains action verbs in multiple levels of the hierarchy. As we move up the hierarchy, the difficulty level of action verb increases. The upper three layers, Analysis, Synthesis and Evaluation demonstrate the verbs with critical thinking. We implement the Bloom's taxonomy as separate part of methodology and restrain its influence on the main sentence similarity methodology. We explain the reason in the following subsection.
\subsubsection{Problem with integrating Bloom's taxonomy with the principal method}
Though Bloom's taxonomy is the suggested standard for designing the course outline, we have found that a considerable number of course drafts differ significantly from the norm. Hence, considering such LOs as well structured and integrating it with the primary methodology violates the purpose. Therefore to use the Bloom's taxonomy, we establish the ``Bloom Index". The Bloom Index represents the learning gap between two learning outcomes according to the verbs in LOs.\newline \\
We start with identifying the action verbs in learning outcomes. Two lists are formed containing the action verbs from each LO respectively. We use Stanford POS Tagger \cite{manning2014stanford} to tag the words and identify action verbs. Each layer in the hierarchy is given a numerical value starting from 1 and going up to 6 as we move up the hierarchy. The absolute distance between the numerical values of layers of verbs yields the distance between two verbs. We use this distance to calculate the index for each pair of the verb. The absolute bloom index for each pair is given by:
\begin{equation}
 absolue\_bloom\_index = \alpha \times distance + \beta
\end{equation}
where $\alpha$ = -0.20 and $\beta$ = 1. 
The \textit{absolue\_bloom\_index} represents the absolute similarity between two verbs according to the Bloom's hierarchy. If both verbs fall into the same category, then they represent the same learning level; hence for such verb pairs it is logical to assign a similarity index which represents maximum similarity. Since, most of the similarity algorithms follow the range from 0 to 1 for the similarity index, we follow the same standard and establish the maximum Bloom similarity as 1. Since the hierarchy is divided into 6 levels uniformaly, and the range for Bloom index is 0 to 1, we set incremental or decremental distance as 0.2. 

We add the absolute bloom indices of all the verb pairs and get the Total Bloom Index. Now, to limit the value of Bloom Index to 0 to 1, we use the total number of comparisons for verb pairs.
Finally, Bloom index is given by:
\begin{equation}
Bloom\_Index= Total\_bloom\_similarity / comparisons
\end{equation}
\subsection{Corpus statistics}
The selection of corpus affects the similarity index by a considerable amount. Learning objectives have some peculiar words. Using a general-purpose corpus does not make justice to such words as the meaning of words differs in general-purpose corpora. No single corpus serves the purpose as the terminologies used in LOs are different for every domain. For example, the terminologies used in Computer Science are different from that of Economics and Chemistry. 
Our similarity algorithm uses Synsets from the WordNet to calculate the semantic similarity between the words. A word can have multiple synsets with different meanings. Hence, it is essential to identify the appropriate synset. \newline
This methodology simulates the use of corpus as a supervised learning model. The corpus is then ``disambiguated", i.e., we find the appropriate sense for each word in the corpus. Identifying sense of the word is part of ``word sense disambiguation" research area. We use `max similarity' algorithm to identify the sense of the words\cite{pedersen2005maximizing}, as implemented in Pywsd, an NLTK based python library\cite{bird2006nltk}.
\begin{equation}
{argmax}_{synset(a)}(\sum_{i}^{n}{{max}_{synset(i)}(sim(i,a))}
\end{equation}
In this stage, we identify the meaning of the word and the synset corresponding to this definition from the WordNet. This information is stored in conjunction with each other to use efficiently for further calculations. The format used is:
Word $\rightarrow$ Synset $\rightarrow$ Meaning of the word \newline
This information also serves as a replica of `Educational Ontologies' synchronous with WordNet.
Then we run a separate thread to establish the frequencies of the synsets and group them according to the meaning. 
The process is repeated everytime the corpus is changed or updated, and new storage is created for every run. In case of rare events, if the disambiguation function fails to tag a word, then we use the statistics from the WordNet. WordNet has the predefined frequency distribution of definition of the words. We use this frequency for the failed words. 

\section{Implementation}
We use previously established Sentence Similarity algorithm which is currently under review elsewhere and modify it as explained in section 3. The database used to implement the proposed methodology is WordNet and statistical information from WordNet. A compiled corpus of the Chemistry domain is used containing learning outcomes, definitions of terminologies and textual contents from books/articles.
\subsection{The Databse - WordNet}
WordNet is a lexical semantic dictionary available for online
and offline use, developed and hosted at Princeton. The WordNet version used for this study is WordNet 3.0 which has 117,000 synonymous sets, Synsets. Synsets of a word represent possible meanings of the word in the context of a sentence.  The central relationship connecting the synsets is the super-subordinate(ISA-HASA) relationship. We use this connection to find the shortest path distance and use this distance to establish similarity between word pairs. 
\begin{table*}
\label{table:1}
\caption{Disambiguated data for LO1}
\begin{tabular}{|c|c|p{7cm}|}
\hline 
subatomic & Synset('subatomic.a.01') & of or relating to constituents of the atom or forces within the atom \\ 
\hline 
composition & Synset('composition.n.03') & a mixture of ingredients \\ 
\hline 
atoms & Synset('atom.n.01') & (physics and chemistry) the smallest component of an element having the chemical properties of the element \\ 
\hline 
ions & Synset('ion.n.01') & a particle that is electrically charged (positive or negative); an atom or molecule or group that has lost or gained one or more electrons \\ 
\hline 
isotopes & Synset('isotope.n.01') & one of two or more atoms with the same atomic number but with different numbers of neutrons \\ 
\hline 
\end{tabular} 
\end{table*}
\setlength{\textfloatsep}{0.1cm}
\subsection{Corpus statistics}
We present a simulation of formation of corpus statistics using a small corpus. Consider following LOs from the corpus.\\
\textit{LO1: Describe the subatomic composition of atoms, ions and isotopes.}\\
\textit{LO2: Calculate spectroscopic quantities in relation to electronic transitions.}\\
\textit{LO3: Write electronic configurations of atoms and ions and relate to the structure of the Periodic Table.}\\
\textit{LO4: An electrical force linking atoms and molecular bonds in chemicals.}\\\\
Table 1 and Table 2 represent the information retrieved from the corpus for LO1 and LO4 respectively. Similarly, all the LOs from the corpus are disambiguated to get the data. We then calculate the frequency distribution of synsets corresponding to words. For instance, the frequency of synset \textit{Synset('atom.n.01')} is 2. Hence, whenever the word \textit{atom} occurs in the LO, the synset considered for semantic similarity calculation will be \textit{Synset('atom.n.01')}. Identically, the statistics are formed for all the synsets and words. Having a well-performing word disambiguation function is crucial to get the precise information from the corpus.

\begin{table*}
\label{table:2}
\caption{Disambiguated data for LO4}
\begin{tabular}{|c|c|p{7.3cm}|}
\hline 
electrical & Synset('electrical.a.01') & relating to or concerned with electricity \\ 
\hline 
force & Synset('power.n.05') & one possessing or exercising power or influence or authority \\ 
\hline 
linking & Synset('connect.v.01') & connect, fasten, or put together two or more pieces \\ 
\hline 
atoms & Synset('atom.n.01') & (physics and chemistry) the smallest component of an element having the chemical properties of the element \\ 
\hline 
bond & Synset('bond.n.01') & an electrical force linking atoms \\ 
\hline 
chemistry& Synset('chemistry.n.02')&the chemical composition and properties of a substance or object\\
\hline
chemical& Synset('chemical.n.01')&material produced by or used in a reaction involving changes in atoms or molecules\\
\hline
\end{tabular} 
\end{table*}

\subsection{Bloom's Taxonomy}
To implement Bloom's Taxonomy, we consider the traditional hierarchical structure. 
We use the verbs listed in Bloom’s Taxonomy of Measurable Verbs\cite{verbs706bloom} arranged in the hierarchy. Each level in the hierarchy is assigned a number starting at 1 with the base `Remembering' and going up to 6 with `Creating'. To tag the verbs in the LOs, we use the Stanford POS tagger \cite{manning2014stanford}.

\subsection{Illustrative Example}
This section explains the working of methodology to calculate the Bloom's Index and the usage of corpus statistics. 
\subsubsection{Calculating Bloom's Index}
Consider following sentences:\\
\textit{S1: Discuss the application of the scientific method to the study of human thinking, development, disorders, therapy, and social processes}\\
\textit{S2: Identify major health informatics applications and develop basic familiarity with healthcare IT products}\\
From S1, we tag one verb, \textit{discuss}.\\
From S2, we tag two verbs,viz., \textit{identify} and \textit{develop}.\\
Here we have 2 comparisons: disucss$\leftrightarrow$identify and develop$\leftrightarrow$discuss. 
Hierarchical distance between disucss$\leftrightarrow$identify is 1, similarly Hierarchical distance between develop$\leftrightarrow$discuss is 3.
Using Eq.(1), we get absolute\_blooms\_index as 0.8 and 0.4 respectively. Now using Eq.(2), we get the Bloom's Index as (0.8+0.4)/2=0.6.

\subsubsection{Corpus Statistics}
Consider two LOs listed in section 4.2.
\textit{LO1: Describe the subatomic composition of atoms, ions and isotopes.}\\
\textit{LO3: Write electronic configurations of atoms and ions and relate to the structure of the Periodic Table.}\\
\begin{table*}
\label{table:3}
\caption{Disambiguated data for LO3}
\begin{tabular}{|c|c|p{7cm}|}
\hline 
electronic & Synset('electronic.a.02') & of or concerned with electrons \\ 
\hline 
configurations & Synset('shape.n.01') & any spatial attributes (especially as defined by outline) \\ 
\hline 
atoms & Synset('atom.n.01') & (physics and chemistry) the smallest component of an element having the chemical properties of the element \\ 
\hline 
atoms & Synset('atom.n.01') & (physics and chemistry) the smallest component of an element having the chemical properties of the element \\ 
\hline 
ions & Synset('ion.n.01') & a particle that is electrically charged (positive or negative); an atom or molecule or group that has lost or gained one or more electrons \\ 
\hline 
structure& Synset('structure.n.03')&the complex composition of knowledge as elements and their combinations\\
\hline
Table& Synset('table.n.05')&a company of people assembled at a table for a meal or game\\
\hline
\end{tabular} 
\end{table*}
\begin{table*}
\label{table:4}
\caption{Similarity between LOs}
\begin{tabular}{|p{5.5cm}|p{5.5cm}|p{2.5cm}|}
\hline
LO1&LO2&Proposed Algorithm Similarity \\
\hline
Acquire knowledge: memorize factual information and laws; assimilate scientific concepts; learn chemical calculations & To predict the physical and chemical properties of organic molecules from structures. & 0.343231930716
\\ \hline
Students will develop both problem solving and critical thinking skills, and they will use these skills to solve problems utilizing chemical principles. & use knowledge of intermolecular forces to predict the physical properties of molecular and extended-network elements and compounds; & 0.295240282004
\\ \hline
apply chemical knowledge to integrate knowledge gained in other courses and to better understand the connections between the various branches of science;&understand and utilize the terminology and concepts of chemistry to acquire and communicate scientific information and to solve basic chemical problems;&0.318542368852
\\ \hline
To become familiar with the structures of organic molecules, especially those found in nature or those with important biological effects;&To predict the physical and chemical properties of organic molecules from structures&0.9405819540\\
\hline
solve problems involving the physical properties of matter in the solid, liquid and gaseous states;&Students will gain an appreciation of the scientific discipline of chemistry and the principles used by chemists to solve complex problems.&0.223101105502
\\ \hline
understand the basis of the unique properties of mixtures and perform related calculations;& memorize factual information and laws; assimilate scientific concepts; learn chemical calculations & 0.289648142927
\\ \hline
apply knowledge of thermochemistry to calculate enthalpy changes associated with chemical and physical processes; & solve problems involving the physical properties of matter in the solid, liquid and gaseous states; & 0.113466429084\\
\hline
Write electronic configurations of atoms and ions and relate to the structure of the Periodic Table.&Describe the subatomic composition of atoms, ions and isotopes.&0.852869346717
\\ \hline
Students will learn and apply the method of inquiry used by chemists to solve chemical problems.&Describe the role of chemists and chemistry in drug design and methods used by chemists.&0.994912072273
\\ \hline
Examine, integrate, and assess any provided or collected chemical data.&Draw scientific conclusions from experimental results or data.&0.900301710749\\
\hline
\end{tabular} 
\end{table*}
Table 1 and Table 3 depicts the words, synsets, and meanings for LO1 and LO3 respectively.\\
From Table 3, considering the meanings of the words, we can conclude that the disambiguation worked fine and we have appropriate synsets for the further calculations; whereas, from Table 3, we conclude that there are some inaccuracies with the words such as \textit{structure} and \textit{table}. The meaning of these words we get after disambiguation is different from their contextual sense in the sentence. The expected meaning of table here is \textit{a tabular array (a set of data arranged in rows and columns)}, and structure is \textit{a structure (the manner of construction of something and the arrangement of its parts)}. \\
Using right set of LOs corresponding to the appropriate domain, we get the synset with the correct meaning. Even while disambiguating the corpus, the disambiguation function can identify inaccurate sense for a word. Using frequency of the sense in corpus deprecates this inaccuracy. 

\section{Experimental Results}
To evaluate the algorithm, we used real Learning Objectives from various course outlines. A survey was conducted and users were asked if they can make a decision based on the resultant semantic similarity and the Bloom Similarity. All the users at least possessed a Bachelors degree. Out of 15 users, 10 users agreed that 75\% or more of the results were useful; 1 user agreed that 65\% or more of the results were useful and 4 users agreed that 55\% or more of the results were useful. 
Table 4 shows the semantic similarity between real-time LOs.

\section{Discussion}
The sentence similarity algorithm used for this methodology achieved good Pearson correlation coefficient of 0.8753 for word similarity concerning the bechmark standard\cite{rubenstein1965contextual} and 0.8794 for sentence similarity with respect to mean human similarity \cite{o2008pilot}. The proposed methodology aims to use this algorithm and make it specific to the Learning Objectives. We use Bloom's Taxonomy to determine the comprehensive similarity between the LOs. We achieve this by establishing relative similarity between verbs. \\
The crucial part of the algorithm is the availability of domain-specific corpus. During this research, we have found no corpus which meets this requirement. So we compiled a small corpus to conduct the study. The contents of the corpus compiled corpus are learning objectives, terminologies and definitions, parts of a book or research belonging to the particular domain. We found that corpus disambiguation works well if we have more apparent words related to that field. This helps the disambiguation function to predict the meaning using the \textit{max\_similarity} algorithm. 
\section{Conclusion}
This paper presented an approach to calculate the semantic similarity between learning objectives using Corpus Statistics and Bloom's taxonomy. The crucial part of the algorithm is the disambiguation of words in the context of their use. Having fewer datapoints may lead to detecting the wrong meaning of the word.   Hence, using a corpus, we make sure that the algorithm always selects the appropriate sense of the word as discussed in the methodology. We use corpus statistics from the disambiguated corpus. The meaning with the highest frequency is considered by the algorithm to find the proper synset from the WordNet. The methodology has been tested on actual learning objectives, and we have achieved very encouraging results. \\
Future work includes expanding the domains and corpus, increasing efficiency of algorithms by using different file structures and forming WordNet-like ontologies for specifically the education domain.

\section{Acknowledgement}
We would like to acknowledge the financial support provided by ONCAT(Ontario Council on Articulation and
Transfer)through Project Number- 2017-17-LU,without their
support this research would have not been possible. We
are also grateful to Salimur Choudhury for his insight on
different aspects of this project; Kyle Robinson for
reviewing and proofreading the paper.

%
%
%
%
%

\setlength{\textfloatsep}{5cm}
\bibliographystyle{plain} 
\bibliography{semantic_similarity}

\end{document}